\documentclass[runningheads]{llncs}
\usepackage{graphicx}
\begin{document}
\title{ \emph{AutoCl}: A Visual Interactive System for Automatic Deep Learning Classifier Recommendation Based on Models Performance.}
\titlerunning{AutoCl}
%
\author{Fuad Ahmed\inst{1} \and
Rubayea Ferdows\inst{2} \and
Md Rafiqul Islam\inst{3} \and
Abu Raihan M. Kamal\inst{1}}
\authorrunning{Ahmed et al.}
%
\institute{Department of Computer Science \& Engineering, Islamic University of Technology (IUT), Bangladesh\\
\email{\{fuadahmed2, raihan.kamal\}@iut-dhaka.edu} \and
Department of Computer Science \& Engineering, International University of Business Agriculture and Technology (IUBAT), Bangladesh\\
\email{rubayea@iubat.edu} \and
Department of Genetics, Genomics, and Informatics, The University of Tennessee Health Science Center (UTHSC), United States \\
\email{rafiqulislam.cse24@gmail.com}}
\maketitle              
\begin{abstract}
Nowadays, deep learning (DL) models being increasingly applied to various fields, people without technical expertise and domain knowledge struggle to find an appropriate model for their task. In this paper, we introduce \emph{AutoCl} a visual interactive recommender system aimed at helping non-experts to adopt an appropriate DL classifier. Our system enables users to compare the performance and behavior of multiple classifiers trained with various hyperparameter setups as well as  automatically recommends a best classifier with appropriate hyperparameter. We compare features of \emph{AutoCl} against several recent AutoML systems and show that it helps non-experts better in choosing DL classifier. Finally, we demonstrate use cases for image classification using publicly available dataset to show the capability of our system.

\keywords{Deep learning  \and Visualization \and Visual Interactive System \and Data Analytics \and Image Classifiers \and Recommendation System.}
\end{abstract}

\section{Introduction} In recent years, we've seen an increase in the use of deep learning (DL) models to support data-driven decision making in various fields. such as speech recognition, computer vision, natural language processing, drug discovery, biomedical sciences, etc. However, despite the effectiveness of DL models, we still have a limited theoretical understanding of these models. Additionally, developing a DL model requires some highly technical expertise and domain knowledge. For example, to improve the performance of DL model, careful selection of model, layer, epoch, optimizer, and batch size is critical.
Classification is one of the most important machine learning problems, hence there is a wide range of classification algorithms are available. However, in most real-world applications, the selection of a classification method for a new dataset or application area is still challenging task.

Numerous automated machine learning (AutoML) methods have been proposed by researchers to fill the gap in human knowledge by automating the DL model developing process.
These systems often have steep learning curves due to manual input and a lack of performance visualization features~\cite{wang2020applying}. As a result, these systems are most of the time inaccessible to the growing number of practitioners who lack the time or background to learn sophisticated tools~\cite{hu2019vizml}. Unlike other common AutoML methods, \emph{AutoCl} is not a black-box optimization process. It can compare and visualize the performance of multiple classifier in class level and recommends best performing classifier with proper hyperparameter.

Most of the exisitng AutoML methods recommends single model as output where \emph{AutoCl} not only recommends a best model but also provides multiple model output as well as gives the users hyperparameter tuning capability. So that it easier to compare the performance of multiple model in class level. 
Therefore, from the existing studies, it is observed that choosing a better model for performing any task is challenging without strong DL expertise and domain knowledge.

\begin{figure}
\centering
\includegraphics[width=12cm,height=9.5cm]{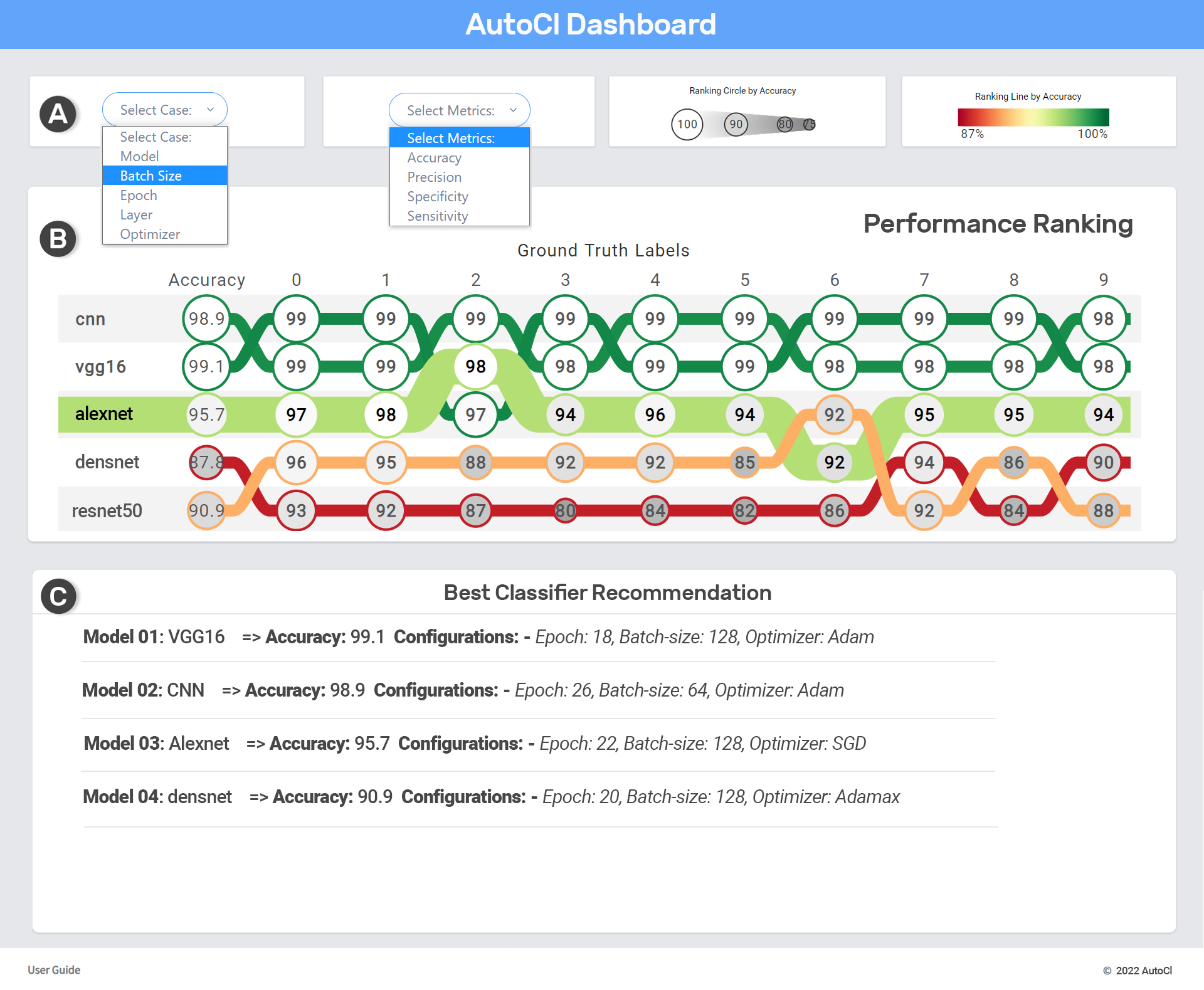}
\caption{The dashboard view of our visual interactive recommender system} \label{fig:dashboard}
\end{figure}


The main objective of this paper is to facilitate non-experts such as researchers and practitioners to choose the most appropriate deep learning classifier for satisfying their requirements. In this paper, we present~\emph{AutoCl}, an interactive deep learning classifier recommendation system as shown in Figure~\ref{fig:dashboard}, where we visualize the performance ranking of multiple DL models and recommend best possible model with best hyperparameter settings. The contributions of this work include:

\begin{itemize}
      
      \item We demonstrate a suite of visualization tools that illustrate the performance of multiple classification algorithm.
      
      \item We show how hyperparameter affects the classifier performance in class level.

      \item Finally the system recommend a top performing classifier with proper hyperparameter.  
      
\end{itemize}

The outline of this paper is organized as follows. In Section~\ref{sec:relatedwork}, we explore the related work, and in Section~\ref{sec:SDD}, we briefly discussed the system design and development to explore the performance of deep learning classifiers. Sections 4 to 6 illustrate the implementation and discussion to evaluate the performance. Finally, in Section~\ref{sec:conclusion}, we provide a conclusion.

\section{Related Work}\label{sec:relatedwork}
Building a deep learning classifier by non-experts is a challenging task, it's also time consuming and error-prone procedure. The challenges motivated many researchers to come up with various AutoML systems to support automatic algorithm selection and Hyper-parameter tuning over the years. In those system they followed statistical approaches, decision trees, neural networks, and other computational intelligence techniques.

In a study Maher and Sakr proposed SmartML~\cite{maher2019smartml} a system for automatic classification algorithm selection and hyperparameter tuning using both meta-learning and Bayesian optimization. Their system limited to work on only 15 machine learning classifier. Auto-sklearn ~\cite{feurer2019auto} is another framework based on scikit-learn for automatic classification and regression task. But it only work for traditional ML algorithms (such as SVM and KNN).

Only a few deep learning framework are proposed for AutoML tasks. Auto-Keras~\cite{jin2019auto} is one of them. It is built based on Keras which is another popular deep learning framework. Auto-Keras focuses on discovering neural network architectures that compete with human-experts. Finally it provides an appropriate deep learning model based on user input data. It also supports multi-modal and multi-task. In addition, very few studies have examined and compared the performance of diffrent AutoML systems~\cite{he2021automl}~\cite{elshawi2019automated}. In general, these studies reveal that there is no obvious winner since there are always trade-offs that must be addressed and optimized based on the context of the problems and the users need.

To the best of our knowledge, this is the first work that automatically recommends best possible deep learning classifier with proper hyperparameter to the non-expert users interactively. It also enables user to tune diffrent hyperparameter settings and compare multiple models performance metrics on class level.

\section{System Design and Development}\label{sec:SDD}

\begin{figure}
\centering
\includegraphics[width=12cm,height=4.5cm]{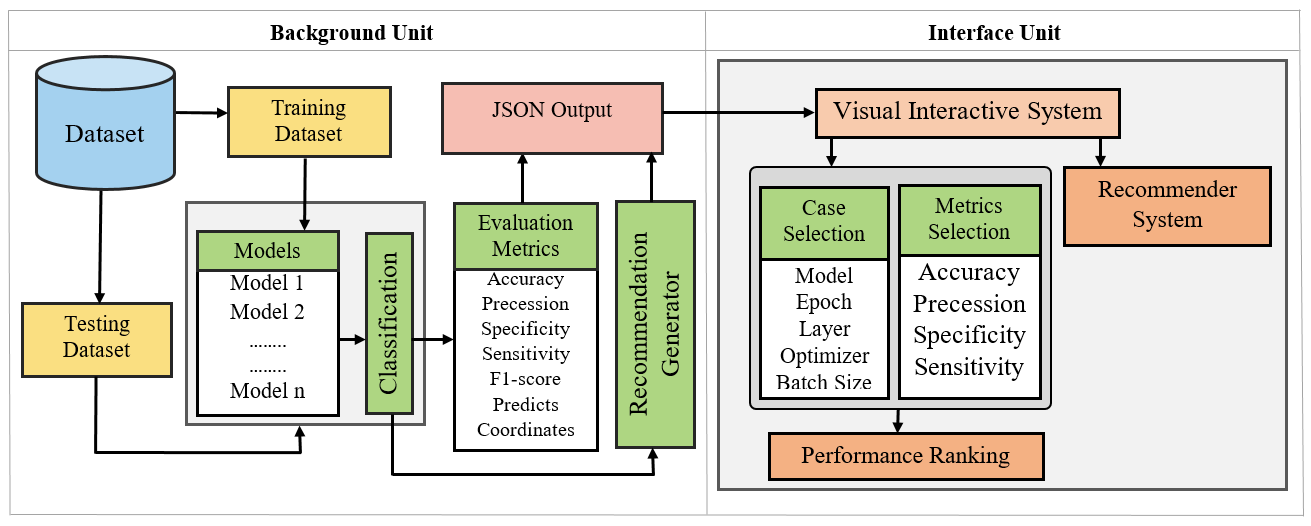}
\caption{System architecture of \emph{AutoCl} System} \label{fig:deepvis}
\end{figure}

We design an interactive recommender system for non-expert users named~\emph{AutoCl} to automatically recommends DL classifier with proper hyperparameter settings. We describe how~\emph{AutoCl} processes data and recmommends best classifer based on user input data. The system can enables user to tune multiple hyperparameter settings to get optimum result. The system architecture of~\emph{AutoCl} shown in Figure~\ref{fig:deepvis} consists of two modules, i.e., (A) background unit and (B) interface unit.

\textbf{A. Background Unit:} The background unit of~\emph{AutoCl} works as a data processor. As the system evaluates multiple models and their metrics, it takes a dataset as an input, and after prepossessing the data, it generates evaluation metrics as output. First, the dataset goes through prepossessing, and it splits into training and testings subsets. It contains file indices, ground truth levels, or class information. The dataset is trained on multiple deep learning classifier and evaluated based on the testing dataset. The background unit calculates high-level statistics about the performance of each model to construct the performance result summary. For automatic recommendation task, AutoKeras~\cite{jin2019auto} library run on multiple classifier and generates a summery of classifiers with best combination of hyperparameter. It do a neural architecture search (NAS) with network morphism based on Bayesian optimization to select a best classifier settings.

Finally, the classifier's performance metrics and recommendation data are saved in a JSON file, and the data goes to the interface unit for visualization. The model's performance data includes model name, hyperparameter settings such as layer structure, epoch number, batch size, optimizer, etc., and metrics such as accuracy, precision, specificity, sensitivity. It also includes list of top ranked classifiers with appropriate hyperparameter for recommendation task.


\textbf{B. Interface Unit:} \emph{AutoCl} interface unit consists of three user interface (UI) modules, and it is responsible for visualizing all the results. As shown in Figure~\ref{fig:dashboard}, modules are visualized as follows: A) This module enables users to choose and tune options such as hyperparameter and metrics. B) A bump chart performance ranking view showing the performance of multiple models in the class level. C) A recommendation view that automatically show the top ranked classifier with suggested hyperparameter settings. Users can adopt best possible classifier for their task suggested by the recommendation module. They can also tune and compare multiple classifier effectively and gather actionable insights determining which classifier to adopt using a combination of different hyperparameter.

\section{Description of AutoCl and Discussion}\label{sec:result}  

\subsection{Dashboard Description} 

In this section, we illustrate how~\emph{AutoCl} analyzes images within models to demonstrate different performances using several DL classifier such as CNN, VGG-16, AlexNet, DenseNet, and ResNet-50. The dashboard of the~\emph{AutoCl}, as shown in Figure.~\ref{fig:dashboard} consists of three significant modules as follows:

\textbf{A. Input Selection:} Users can filter hyperparameter and accuracy metrics using the~\emph{AutoCl} option controls. They can interactively explore and compare models based on performance ranking and ground-truth labels as shown in Figure.~\ref{fig:dashboard}(A). Also, users can further adjust hyperparameters such as batch size, epoch, layer, and optimizer to see what parameter combination has higher performance results.

\textbf{B. Performance Ranking:} The performance ranking module allows DL practitioners to evaluate the performance of multiple models simultaneously. They can assess both class-level performances as well as individual performance metrics of any class. In Figure~\ref{fig:dashboard}(B), we develop a bump chart to visualize multiclass and multimodel performance at the same time. In the chart, each model is represented by a ranking line in which columns represent the ground truth level of class. Each class denotes by a circle, and it shows the performance metric value inside it. Circle size increases or decreases depending on the measured value of that class. The models in the ranking line represented by red to green color aligned with their performance metric value. Once a user clicks any model in the chart, that model's ranking line becomes bold, and it is easier to look at which class has poor performance. We can see that both CNN and VGG-16 have a higher average performance and stable prediction for the entire class whereas DenseNet and ResNet-50 have poor performance.

\textbf{C. Recommendation System:} In the Automatic recommender view, user can see a list of classifier shown based on accuracy and other metrics. The classifiers also have set of hyperparameter settings with each model name shown in Figure.~\ref{fig:dashboard}(C). This classifier rank generated based on each classifier performance.

\section{Usage Scenario: Image Classification}\label{sec:usage}
In this section, we show how~\emph{AutoCl} can effectively support
non-experts to choose best deep learning classifier for their work.
Specifically, we demonstrate how multiple classifier performance comparison and tuning hyperparameter can support their exploration.

John, a student who is interested in image classification problem. For a study he need to select a image classifier. He wants to know, how an image classifier model classify images. He has done some data visualization before, but he doesn't have much deep learning expertise. He finds CIFAR-10 dataset, which is publicly available~\cite{krizhevsky2009learning}. It is a multiclass dataset containing 60000 32x32 color images. It has ten (10) classes: airplane, automobile, bird, cat, deer, dog, frog, horse, ship, and truck. Each class has exactly 6,000 images. There are 50,000 training images and 10,000 test images in the dataset. 

He loads the dataset in~\emph{AutoCl} system and specify few classifier (CNN, VGG-16, AlexNet, DenseNet, and ResNet-50) and mentions the classes he want to predict. Once he run the~\emph{AutoCl} Background Unit, the system process the data and calculate necessary metrics and recommender data. Then the data shows in Interface Unit. He can see that the~\emph{AutoCl} recommender view suggest him top performing classifier with necessary hyperparameter. CNN performs better where ResNet50 performs poorly.

He also want to know which classifier provide better performance on class level. From the~\emph{AutoCl} Option Selection and Performance Ranking view~\ref{fig:dashboard}(A-B), he sets different hyperparameter and observe the performance metrics such as accuracy precision instantly. He selects the epoch from option settings and discover that the performance varies for different epoch number. The check all the options settings and interactively discover and learnt how the classifier classifies the image in class level. After using the~\emph{AutoCl} dashboard, finally he come to a decision that CNN is the most performing classifier for his task on given CIFAR-10 dataset.

\begin{table}

\centering
\caption{Feature Comparison between~\emph{AutoCl} and few other AutoML Framework}\label{tab:compare}
\begin{tabular}{|p{1cm}|p{5.2cm}|p{1.2cm}|p{1.4cm}|p{1cm}|p{1.5cm}|}
\hline

No. & Features of dashboard & Smart ML & Auto-sklearn & Auto-Keras & \textbf{\textbf{AutoCl}}\\
\hline\hline
F1 & Multiple Model Compare & No & No & No  & Yes \\
\hline
F2 & Performance ranking & No & No & No  & Yes \\
\hline
F3 & Support Deep Learning & No & No & Yes  & Yes \\
\hline
F4 & Visual Interpretation & No & No & No  & Yes\\
\hline
F5 & Hyperparameter Tuning & No & No & Yes  & Yes \\
\hline
F6 & Automated process & Yes & No & No  & Yes \\
\hline
\end{tabular}\\

\end{table}

\section{Evaluation}\label{sec:evalation}

We conduct a two-stage evaluation study to assess the potential usability and usefulness of our system. In the first stage, we demonstrated a usage scenario as described in section~\ref{sec:usage}. In the second stage,~\emph{AutoCl} is compared with three different recommendation system as shown in Table~\ref{tab:compare} where (Yes) and (No) indicates the presence and absence of the feature selections~\cite{maher2019smartml}, ~\cite{feurer2019auto}, ~\cite{jin2019auto}. In short, ~\emph{AutoCl} is useful for recommending classifier and evaluating performance at the class levels to compare multiple classifier effectively. 

\section{Conclusion}\label{sec:conclusion}

In this paper, we presented~\emph{AutoCl}, an interactive recommender system to suggest best performing classifier to non-expert users. Although there are  several cloud based and opensource AutoML system exits, those are either costly or not feature rich. Our system is helpful for non-experts to effectively select deep learning classifier for their work. It help them interactively tune hyperparameter and suggest best performing classifier. In the future, we will improve the system to be used as a visual interactive DL tool to update and learn additional models at the instance level with a higher accuracy rate. We will also incorporate our system for other deep learning task such as prediction, natural language processing etc.
%
%
%
\bibliographystyle{splncs04}
\bibliography{Reference.bib}
\end{document}